# Balancing Exploration and Exploitation by an Elitist Ant System with Exponential Pheromone Deposition Rule

Ayan Acharya[1], Deepyaman Maiti[2], Aritra Banerjee[3], Amit Konar[4]
[1,2,3,4]Department of Electronics and Telecommunication Engineering,
Jadavpur University, Kolkata: 700032
[1]masterayan@gmail.com, [2]deepyamanmaiti@gmail.com, [3]aritraetece@gmail.com, [4]konaramit@yahoo.co.in

*Abstract*— The paper presents an exponential pheromone deposition rule to modify the basic ant system algorithm which employs constant deposition rule. A stability analysis using differential equation is carried out to find out the values of parameters that make the ant system dynamics stable for both kinds of deposition rule. A roadmap of connected cities is chosen as the problem environment where the shortest route between two given cities is required to be discovered. Simulations performed with both forms of deposition approach using Elitist Ant System model reveal that the exponential deposition approach outperforms the classical one by a large extent. Exhaustive experiments are also carried out to find out the optimum setting of different controlling parameters for exponential deposition approach and an empirical relationship between the major controlling parameters of the algorithm and some features of problem environment.

## I. INTRODUCTION

**Ant Colony Optimization (ACO)** is a paradigm for designing metaheuristic algorithms for combinatorial optimization problems. While roaming from food sources to the nest and vice versa, ants deposit on the ground a substance called *pheromone*. Ants can smell pheromone and choose, in probability, paths marked by stronger pheromone concentrations. Hence, the pheromone trail allows the ants to find their way back to the food source or to the nest. ACO algorithm simulates this behavior of ant colony to solve difficult **NP** hard optimization problems.

**Ant System (AS)** is the earliest form of ant colony optimization algorithm that has been modified by numerous researchers till date. **Elitist Ant System (EAS)** model is one such improved model of the primary version of ant system. Our paper extends the **AS** model by introducing an exponential pheromone deposition approach, contrary to the uniform deposition approach used in classical **AS** algorithms. We attempt to solve the deterministic **AS** dynamics using differential equation. This novel analysis helps in determining the range of parameters in the exponential pheromone deposition rule to confirm stability in pheromone trails. The deterministic solution does not violate the stochastic nature of the **AS** because a segment of trajectory here is always selected probabilistically.

Our previous work [8] was based on stability analysis using difference equation. In this paper, we have employed differential equations which not only characterize the system more precisely but also are more popular than difference equations. The previous paper, with experiments performed over TSP instances, could not at all highlight the philosophy of the non uniform deposition rule. This paper presents sufficient simulation backup to establish the proposed algorithm's superiority over the traditional one. Problem environment is also chosen very cleverly to emphasize the efficacy of the proposed algorithm. Exhaustive experimentations also help find out the suitable values of parameter for which the proposed algorithm works best and from these results we try to ascertain an algebraic relationship between the parameter set of the algorithm and feature set of the problem environment.

The paper is structured in 6 sections. In section II, a brief introduction of **AS** and **EAS** are provided. We formulate a scheme for the general solution of the Ant System in section III. Stability analysis with complete solution for different pheromone deposition rules is undertaken in section IV. Performance analyses of the proposed and classical AS are compared in section V on elitist model. Finally, the conclusions are listed in section VI.

## II. ANT SYSTEM AND ELITIST ANT SYSTEM

The theory of ant system can best be explained in the context of Travelling Salesperson Problem (TSP)([6]). The basic ACO algorithm for TSP can be described as follows:
**procedure** ACO algorithm for TSPs
➢ Set parameters, initialize pheromone and ants' memory
   **while** (termination condition not met)
➢ Construct Solution
➢ Apply Local Search ( optional)
➢ Best Tour check
➢ Update Trails
   **end**
**end** ACO algorithm for TSPs
  **Ant System (AS)** ([1],[2],[3]) basically consists of two levels:

1. **Initialization: 1.** Any initial parameters are loaded. **2.** Edges are set with an initial pheromone value. **3.** Each ant is individually placed on a random city.
2. **Main Loop:**
- **Construct Solution**
  Each ant constructs a tour by successively applying the probabilistic choice function:

$$P_i^k(j) = \begin{cases} (\tau_{ij}^\alpha)(\eta_{ij}^\beta) / \sum_{k: k \in N_i^k} (\tau_{ik}^\alpha)(\eta_{ik}^\beta) & \text{if } q < q_0 \\ 1 \text{ if } (\tau_{ij}^\alpha)(\eta_{ij}^\beta) = \max\{(\tau_{ik}^\alpha)(\eta_{ik}^\beta) : k \in N_i^k\} & \text{with } q > q_0 \\ 0 \text{ if } (\tau_{ij}^\alpha)(\eta_{ij}^\beta) \neq \max\{(\tau_{ik}^\alpha)(\eta_{ik}^\beta) : k \in N_i^k\} & \text{with } q > q_0 \end{cases} \quad (1)$$

where $P_i^k(j)$ is the probability of selecting node $j$ after node $i$ for ant $k$. A node $j \in N_i^k$ ($N_i^k$ being the neighborhood of ant $k$ when it is at node $i$) if $j$ is not already visited. $\eta_{ik}$ is the visibility information generally taken as the inverse of the length of link $(i,k)$, is the pheromone concentration associated with the link $(i,k)$. $q_0$ is a pseudo random factor deliberately introduced for path exploration and α, β are the weights for pheromone concentration and visibility.

- **Best Tour check:** Calculate the lengths of the ants' tours and compare with best tour length so far. If there is an improvement, update it.
- **Update Trails: 1.** Evaporate a fixed proportion of the pheromone on each edge. **2.** For each ant perform the **'Ant Cycle'** ([3]) pheromone update.

First improvement over **AS** was proposed as the **Elitist Ant System (EAS)** strategy ([2],[3],[9]) in which additional reinforcement is provided to the best solution found from the start of the algorithm. Now, let $i$ and $j$ be two successive nodes, on the tour of an ant and $\tau_{ij}(t)$ be the pheromone concentration created by the ant at time t and associated with the edge of the graph joining the nodes $i$ and $j$.

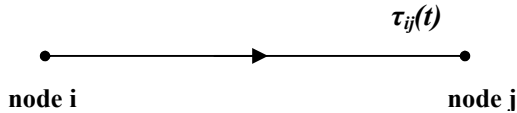

**Fig. 1: Defining $\tau_{ij}(t)$**

Let $\rho > 0$ be the pheromone evaporation rate, and $\Delta\tau_{ij}^k(t)$ be the pheromone deposited by ant k at time t. The basic pheromone updating rule in **AS** is then given by,

$$\tau_{ij}(t) = (1-\rho)\tau_{ij}(t-1) + \sum_{k=1}^{m} \Delta\tau_{ij}^k(t) \quad (2)$$

In Elitist model, a special preference is given to the best path found so far. Thus the pheromone update rule for the best so far tour is:

$$\tau_{ij}(t) = (1-\rho)\tau_{ij}(t-1) + \sum_{k=1}^{m} \Delta\tau_{ij}^k(t) + e\Delta\tau_{ij}^{bs} \quad (3)$$

where $\Delta\tau_{ij}^k$ is the amount of pheromone deposited by ant $k$ on the arcs it has visited and is defined as follows:

$$\Delta\tau_{ij}^k = \begin{cases} 1/C_k, & \text{if arc } (i,j) \text{ belongs to } T^k \\ 0, & \text{otherwise} \end{cases}, C_k \text{ being the length of}$$

the tour $T^k$ constructed by $k^{th}$ ant. e is a parameter that defines the weight given to the best-so-far tour $T^{bs}$ with tour length $C^{bs}$. $\Delta\tau_{ij}^{bs}$ in (3) is defined as

$$\Delta\tau_{ij}^{bs} = \begin{cases} 1/C^{bs}, & \text{if arc } (i,j) \text{ belongs to } T^{bs} \\ 0, & \text{otherwise} \end{cases}. \text{ A suitable choice of}$$

parameter **e** allows **EAS** to find better tour in a smaller number of iterations compared to **AS**.

### III. DETERMINISTIC FRAMEWORK FOR SOLUTION OF BASIC ANT SYSTEM DYNAMICS

Now, from (2), $\tau_{ij}(t) - \tau_{ij}(t-1) = -\rho\tau_{ij}(t-1) + \sum_{k=1}^{m} \Delta\tau_{ij}^k(t)$

$$\Rightarrow \frac{d\tau_{ij}}{dt} = -\rho\tau_{ij} + \sum_{k=1}^{m} \Delta\tau_{ij}^k(t) \therefore (D+\rho)\tau_{ij}(t) = \sum_{k=1}^{m} \Delta\tau_{ij}^k(t+1) \quad (4)$$

where we define $\frac{d\tau_{ij}}{dt} = \tau_{ij}(t) - \tau_{ij}(t-1) = D\tau_{ij}$

Evidently, (4) gives the solution for the ant dynamics. Now, to solve (4), we have to separate the complimentary function and the particular integral. We now consider two different forms of $\Delta\tau_{ij}^k(t)$ and try to determine the complete solution of $\tau_{ij}(t)$.

**Evaluation of Complimentary Function:**

The complimentary function of (4) is obtained by setting $\sum_{k=1}^{m} \Delta\tau_{ij}^k(t)$ to zero. This gives only the transient behavior of the ant system dynamics. Therefore, from (4), $(D+\rho)\tau_{ij} = 0, \Rightarrow D = -\rho$

Thus, the transient behavior of the Ant System is given by

$$\text{CF: } \tau_{ij}(t) = Ae^{-\rho t} \quad (5)$$

where $A$ is a constant which is to be determined from initial condition.

**Evaluation of Particular Integral for Both Forms of Deposition Rule:**

The steady state solution of the ant system dynamics is obtained by computing particular integral of (4). This is given by,

$$\tau_{ij} = \frac{1}{D+\rho} \sum_{k=1}^{m} \Delta\tau_{ij}^k(t+1) \quad (6)$$

**Case I:** When $\Delta\tau_{ij}^k(t) = C_k$, we obtain from (6)

$$PI = \frac{1}{D+\rho} \sum_{k=1}^{m} C_k = \frac{1}{\rho}(1 + D/\rho)^{-1} \sum_{k=1}^{m} C_k$$

$$= \frac{1}{\rho}(1 - \frac{D}{\rho} + \frac{D^2}{\rho^2} - \dots) \sum_{k=1}^{m} C_k = \frac{1}{\rho}(1) \sum_{k=1}^{m} C_k = \sum_{k=1}^{m} C_k/\rho \quad (7)$$

**Case II:** When $\Delta\tau_{ij}^k(t) = C_k(1 - e^{-t/T})$, we obtain from (6),

$$PI = \frac{1}{D+\rho}\sum_{k=1}^{m}C_k(1-e^{-(t+1)/T}) = \frac{1}{D+\rho}\sum_{k=1}^{m}C_k - \frac{1}{D+\rho}\sum_{k=1}^{m}C_k e^{-(t+1)/T}$$

$$= \sum_{k=1}^{m}C_k/\rho - \frac{1}{D+\rho}\sum_{k=1}^{m}C_k e^{-(t+1)/T} = \sum_{k=1}^{m}C_k/\rho - \sum_{k=1}^{m}C_k e^{-(t+1)/T}/(\rho-\frac{1}{T}) \quad (8)$$

## IV. STABILITY ANALYSIS OF ANT SYSTEM DYNAMICS WITH COMPLETE SOLUTION

In this section, we obtain the closed form solution of the ant system dynamics for determining the condition for stability of the dynamics.

**Case I:** For constant deposition rule, the complete solution can be obtained by adding CF and PI from (5) and (7) respectively and is given by,

$\tau_{ij}(t) = Ae^{-\rho t} + \sum_{k=1}^{m}C_k/\rho$. At t=0,

$\tau_{ij}(0) = A + \sum_{k=1}^{m}C_k/\rho$, $\Rightarrow A = \tau_{ij}(0) - \sum_{k=1}^{m}C_k/\rho$

Therefore, the complete solution is,

$$\tau_{ij}(t) = [\tau_{ij}(0) - \sum_{k=1}^{m}C_k/\rho]\, e^{-\rho t} + \sum_{k=1}^{m}C_k/\rho \quad (9)$$

It follows from (9) that the system is stable for $\rho>0$ and converges to steady state value $\sum_{k=1}^{m}C_k/\rho$ as time increases. The plot below supports the above observation.

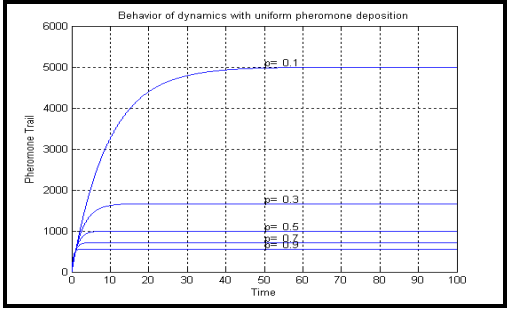

**Figure 2:** $\tau_{ij}(t)$ versus t for constant pheromone deposition

**Case II:** For exponentially increasing pheromone deposition, the complete solution is,

$$\tau_{ij}(t) = Ae^{-\rho t} + \sum_{k=1}^{m}C_k/\rho - \sum_{k=1}^{m}C_k e^{-(t+1)/T}/(\rho-\frac{1}{T})$$

Now, at t=0, $\tau_{ij}(0) = A + \sum_{k=1}^{m}C_k/\rho - \sum_{k=1}^{m}C_k e^{-1/T}/(\rho-\frac{1}{T})$

Therefore, with initial condition incorporated, the overall solution is given by,

$$\tau_{ij}(t) = \tau_{ij}(0)\, e^{-\rho t} + \sum_{k=1}^{m}\frac{C_k}{\rho}(1-e^{-\rho t}) + \sum_{k=1}^{m}\frac{C_k e^{-(\rho t+1/T)}}{(\rho-\frac{1}{T})}(1-e^{-[(1/T)-\rho]t}) \quad (10)$$

Clearly, the system is stable for positive values of ρ and T and converges to $\sum_{k=1}^{m}C_k/\rho$ in its steady state.

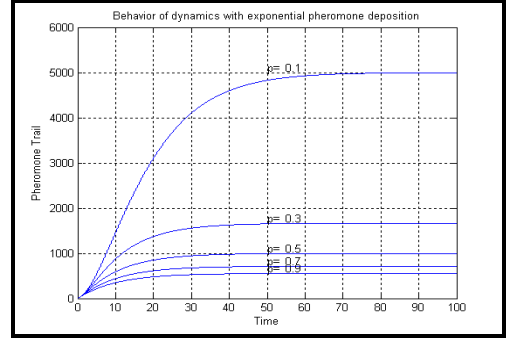

**Figure 3:** $\tau_{ij}(t)$ versus t for exponential pheromone deposition with T=10

A uniform pheromone deposition by an ant cannot ensure subsequent ants to follow the same trajectory. However, an exponentially increasing time function ensures that subsequent ants close enough to a previously selected trial solution will follow the trajectory, as it can examine gradually thicker deposition of pheromones over the trajectory. Naturally, ***deception probability*** ([4]) being less, convergence time should improve.

## V. SIMULATION RESULTS

The **EAS** model is considered here to study the performance of the ant system algorithm with exponential deposition rule. As a problem environment, we take a network of connected cities where the shortest route between two given cities is to be determined. Ants begin their tour at the starting city and terminate their journey at the destination city and decide its next position at each intermediate step by a probability based selection approach as given in (1). Interpretation of different terms in (1) is exactly same as in context of TSP except the term $\eta_{ik}$ which is defined here as $\eta_{ik}=1/(|d_{ik}|+|d_{kg}|)$ where $d_{ik}$ is defined as the distance between the cities *i* and *k* and $d_{kg}$ specifies the distance between cities *k* and *g*, *g* being the destination city. **α, β** are the weights for pheromone concentration and visibility as usual. Ants also stop moving if they find a dead end.

In constant pheromone approach, deposition of excess pheromone in all links of a path is kept constant. But in our approach, pheromone deposition is gradually increased in the links near the destination city. It implies that the links lying closer to the destination city receive more pheromone compared to those near the starting city.

We divide the simulation strategy in two different levels. In the first level, we run the two competitive algorithms on 10 different city distributions and estimate the range of values of parameters of the proposed algorithm for which it performs best and outperforms its classical counterpart by largest extent. In following section, we tabulate results for only 3 different distributions owing to space constraint.

*A.* **Level I Results:**

**Results for Environment I:**

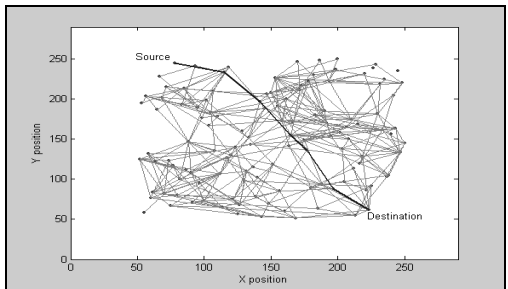

**Fig 4 : Roadmap for 120 City Distribution**

The first experiment is conducted with 120 cities. 20 ants are employed to move through the graph for 100 iterations. We vary both α and β over the range 0.5 to 5.0 in steps of 0.5 and best optimum path length was obtained for α=1.5, β=4.0. Length of the best path found with above parameter setting almost matches with the theoretical minima as obtained by applying Dijkstra's algorithm([16]). That path is marked by bold black line in figure 4. Table 1 provides the variation of convergence time (number of iterations required for optimum path length to converge) with the variation of α and β. The convergence time with α=1.5 and β=4.0 is near to optimum value (16) which signifies that for the above roadmap α=1.5, β=4.0 is the optimum parameter setting which not only produces optimum solution but also in fairly optimum number of iterations. A 3D plot of optimum path length for varying α, β is provided in figure 5.

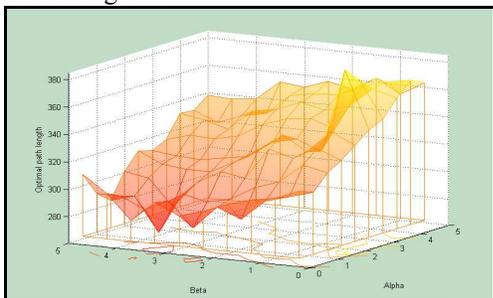

**Fig 5 : Variation of Optimum Pathlength with α and β**

**Table 1: Variation of convergence time with α and β**

| α / β | 0.5 | 1.0 | 1.5 | 2.0 | 2.5 | 3.0 | 3.5 | 4.0 | 4.5 | 5.0 |
|---|---|---|---|---|---|---|---|---|---|---|
| 0.5 | 37 | 40 | 43 | 44 | 46 | 47 | 52 | 60 | 55 | 58 |
| 1.0 | 35 | 38 | 44 | 47 | 49 | 52 | 53 | 57 | 61 | 64 |
| 1.5 | 38 | 32 | 35 | 38 | 40 | 44 | 52 | 53 | 55 | 60 |
| 2.0 | 34 | 31 | 32 | 31 | 38 | 40 | 37 | 44 | 47 | 52 |
| 2.5 | 32 | 28 | 26 | 32 | 35 | 38 | 32 | 46 | 43 | 55 |
| 3.0 | 29 | 30 | 23 | 29 | 40 | 35 | 31 | 41 | 44 | 50 |
| 3.5 | 28 | 23 | 20 | 26 | 28 | 32 | 38 | 47 | 50 | 53 |
| 4.0 | 25 | 16 | 19 | 22 | 29 | 31 | 40 | 43 | 47 | 55 |
| 4.5 | 29 | 23 | 25 | 26 | 32 | 37 | 32 | 41 | 53 | 59 |
| 5.0 | 32 | 34 | 31 | 37 | 40 | 43 | 44 | 46 | 47 | 52 |

In all simulations above, we assume T=10. This value of T is guessed from the number of links required to move from source city to destination city which, for most optimal solutions, lies between 12 and 15. Therefore, T=10 is a reasonable approximation as far as the philosophy of exponential deposition rule is concerned. Still further tuning of T is necessary and hence experiment with varying value of T in the neighborhood of its estimated value is conducted with the optimal setting of parameters **α, β** i.e. α=1.5 and β=4.0. The result is presented in table 2.

**Table 2: Variation of convergence time with T**

| T | Convergence Time | T | Convergence Time |
|---|---|---|---|
| 7.0 | 21 | 10.5 | 19 |
| 7.5 | 19 | 11.0 | 20 |
| 8.0 | 18 | 11.5 | 19 |
| 8.5 | 17 | 12.0 | 21 |
| 9.0 | 18 | 12.5 | 20 |
| 9.5 | 16 | 13.0 | 22 |
| 10.0 | 19 | | |

A comparative study of the two competitive algorithms is carried out next with optimum parameter settings. The plot depicts the superiority of the proposed method in terms of both solution quality and convergence time. In figure (6), the red graph shows the iteration-best paths for exponential deposition rule and the blue graph shows the same for constant deposition rule. The line marked green shows the theoretical minimum path-length between the source and destination cities. For simulating the constant deposition algorithm α=1 and β=2 (as suggested in [3]) are used. *e* is set to number of nodes present in case of TSP ([3]). The closed path found by the ants includes all the cities in TSP. But in our problem, the optimum paths found by the ants do not consist of more than 15 cities. The parameter *e*, in our problem, is therefore set at 15 for simulating both forms of pheromone deposition rule.

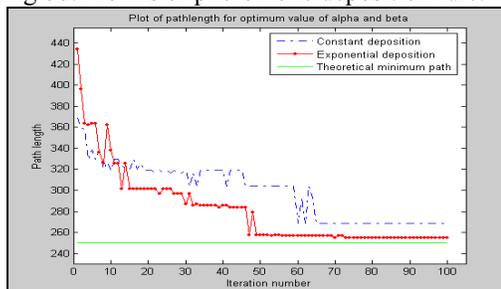

**Fig 6 : Comaparative Study of algorithms**

**Results for Environment II:**

Environment II is slightly more complicated distribution with 180 cities. Experiments conducted led to the optimum parameter setting **α=1.0, β=3.5, T=11.0**.

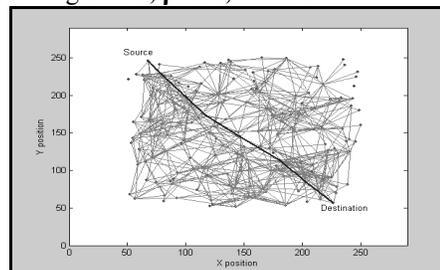

**Fig 7: Roadmap for 180 City Distribution**

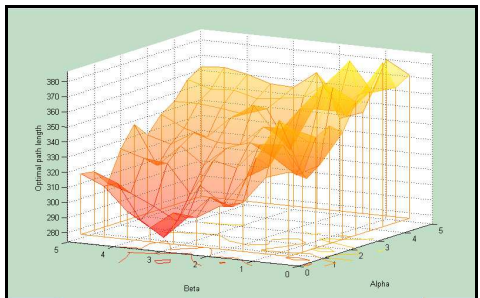

**Fig 8 : Variation of Optimum Pathlength with α and β**

**Table 3: Variation of convergence time with α and β**

| α\β | 0.5 | 1.0 | 1.5 | 2.0 | 2.5 | 3.0 | 3.5 | 4.0 | 4.5 | 5.0 |
|---|---|---|---|---|---|---|---|---|---|---|
| 0.5 | 51 | 48 | 57 | 58 | 58 | 62 | 68 | 76 | 66 | 73 |
| 1.0 | 48 | 52 | 60 | 60 | 63 | 65 | 64 | 67 | 74 | 73 |
| 1.5 | 52 | 43 | 52 | 51 | 48 | 52 | 68 | 62 | 70 | 74 |
| 2.0 | 44 | 43 | 48 | 44 | 51 | 48 | 44 | 55 | 58 | 70 |
| 2.5 | 45 | 40 | 32 | 45 | 48 | 55 | 50 | 58 | 56 | 71 |
| 3.0 | 43 | 42 | 41 | 45 | 56 | 46 | 43 | 54 | 58 | 60 |
| 3.5 | 42 | 28 | 34 | 40 | 38 | 46 | 54 | 57 | 61 | 65 |
| 4.0 | 36 | 29 | 30 | 33 | 40 | 48 | 58 | 52 | 62 | 64 |
| 4.5 | 40 | 35 | 34 | 40 | 45 | 52 | 41 | 52 | 63 | 66 |
| 5.0 | 48 | 48 | 45 | 50 | 52 | 52 | 60 | 64 | 62 | 64 |

**Table 4 : Variation of convergence time with T**

| T | Convergence Time | T | Convergence Time |
|---|---|---|---|
| 7.0 | 33 | 10.5 | 29 |
| 7.5 | 30 | 11.0 | 25 |
| 8.0 | 31 | 11.5 | 26 |
| 8.5 | 32 | 12.0 | 28 |
| 9.0 | 30 | 12.5 | 29 |
| 9.5 | 29 | 13.0 | 28 |
| 10.0 | 28 | | |

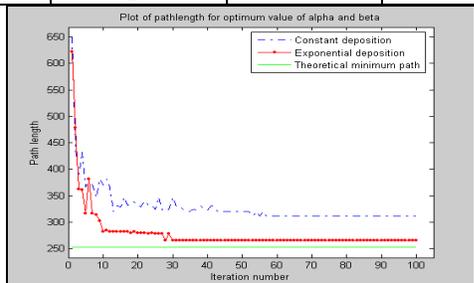

**Fig 9 : Comaparative Study of algorithms**

**Results for Environment III:**

Figure 10 shows a roadmap of 240 city distribution, an extremely complicated graph. Optimum performance is achieved at **α=0.5,β=3.5 and T=12.0**.

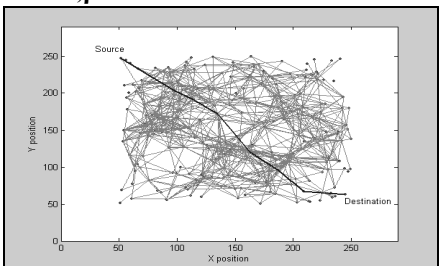

**Fig 10: Roadmap for 240 City Distribution**

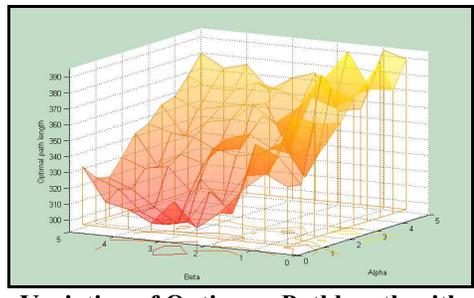

**Fig 11 : Variation of Optimum Pathlength with α and β**

**Table 5: Variation of convergence time with α and β**

| α\β | 0.5 | 1.0 | 1.5 | 2.0 | 2.5 | 3.0 | 3.5 | 4.0 | 4.5 | 5.0 |
|---|---|---|---|---|---|---|---|---|---|---|
| 0.5 | 60 | 67 | 64 | 68 | 71 | 76 | 77 | 87 | 73 | 86 |
| 1.0 | 64 | 57 | 69 | 70 | 68 | 73 | 73 | 80 | 86 | 83 |
| 1.5 | 66 | 52 | 57 | 64 | 62 | 68 | 80 | 78 | 80 | 85 |
| 2.0 | 55 | 57 | 57 | 50 | 65 | 69 | 59 | 70 | 71 | 71 |
| 2.5 | 55 | 48 | 51 | 59 | 60 | 65 | 56 | 71 | 69 | 86 |
| 3.0 | 46 | 55 | 53 | 54 | 62 | 59 | 58 | 64 | 71 | 77 |
| 3.5 | 35 | 47 | 49 | 55 | 47 | 60 | 60 | 77 | 79 | 76 |
| 4.0 | 44 | 43 | 55 | 53 | 49 | 50 | 61 | 61 | 70 | 79 |
| 4.5 | 50 | 40 | 54 | 47 | 58 | 56 | 58 | 65 | 78 | 84 |
| 5.0 | 50 | 55 | 58 | 65 | 66 | 68 | 70 | 69 | 78 | 74 |

**Table 6: Variation of convergence time with T**

| T | Convergence Time | T | Convergence Time |
|---|---|---|---|
| 7.0 | 37 | 10.5 | 35 |
| 7.5 | 35 | 11.0 | 34 |
| 8.0 | 36 | 11.5 | 34 |
| 8.5 | 38 | 12.0 | 32 |
| 9.0 | 37 | 12.5 | 35 |
| 9.5 | 35 | 13.0 | 34 |
| 10.0 | 35 | | |

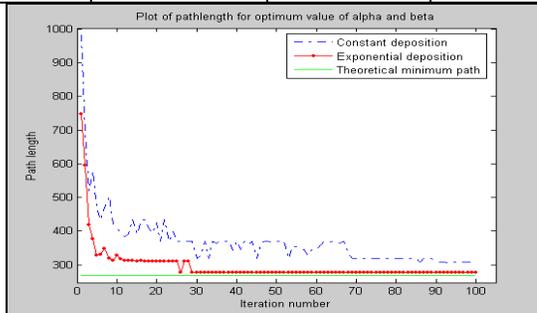

**Fig 12: Comaparative Study of algorithms**

*B.* **Level II Results:**

Experiments performed above reveal that the proposed algorithm performs best for **α** lying between 0.5 and 1.5 and **β** lying between 3.5 and 4.0, no matter how complex the environment is. In secondary level of our simulation strategy, we vary **α** and **β** over the above mentioned range in steps of 0.1 and try to estimate their relation with two features of problem environment: i) the node density and ii) standard deviation of lengths of smallest arc associated with each node. We performed experiments on roadmaps with 120,140,160,180,200,220 and 240 number of cities. For each of above roadmaps, we chose seven different distributions and recorded the values of **α** and **β** for best performance. **Table Curve 3D V4.0**, a curve fitting tool, was then employed to fit a curve through 49 data points for each of **α** and **β** and obtain an algebraic relation between **α** or **β** and the features of problem environment. The results are displayed in the following two figures (fig 13 and 14) along with the approximated equations establishing the relation between two sets of parameters. This exhaustive experimentation allows determination of optimum values of **α** and **β** when the features of problem environment are known in advance.

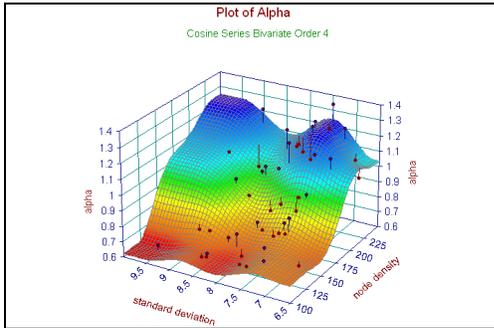

**Fig 13: Plot of α**
**Function: Cosine Series Bivariate Order 4**
[ x": x scaled 0 to $\pi$, y": y scaled 0 to $\pi$;
x ≡ number of nodes in 200 sq unit area, y ≡ standard deviation]
α=a+bcos(x")+ccos(y")+dcos(2x")+ecos(x")cos(y")+ fcos(2y")+ gcos(3x")+hcos(2x")cos(y")+icos(x")cos(2y") +jcos(3y")+ kcos(4x")+lcos(3x")cos(y")+mcos(2x")cos(2y") + ncos(x")cos(3y")+ocos(4y")

**Co-efficient values:**
a=0.935, b=-0.237, c=-0.020, d=-0.011, e=-0.028,
f=0.028, g=-0.002, h=0.027, i=0.0006, j=-0.039
k=-0.006, l=0.056, m=-0.022, n=0.047, o=-0.020

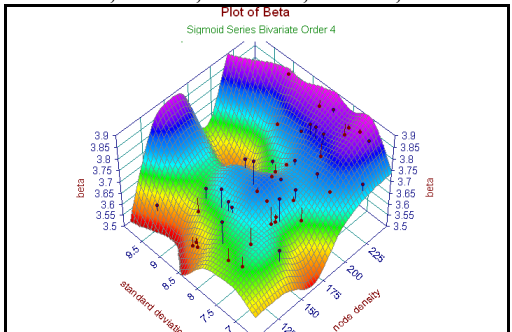

**Fig 14: Plot of β**
**Function: Sigmoid Series Bivariate Order 4**
[ x': x scaled -1 to +1, y': y scaled -1 to +1
$S_i$=2..n(x')=-1+2/(1+exp(-(x'+1-(i-1)*(2/n))/0.12)), $S_1$(x')= x']
β=a+b$S_1$(x')+c$S_1$(y')+d$S_2$(x')+e$S_1$(x')$S_1$(y')+f$S_2$(y')+ g$S_3$(x')+ h$S_2$(x')$S_1$(y')+i$S_1$(x')$S_2$(y')+j$S_3$(y')+k$S_4$(x')+ l$S_3$(x')$S_1$(y') + m$S_2$(x')$S_2$(y')+n$S_1$(x')$S_3$(y')+o$S_4$(y')

**Co-efficient values:**
a=3.742, b=0.323, c=0.422, d=-0.090, e=0.414
f=-0.124, g=-0.105, h=-0.12, i=-0.131, j=-0.111
k=0.019, l=-0.196, m=0.100, n=0.007, o=-0.139

## VI. CONCLUSIONS AND SCOPE OF FUTURE WORK

The paper presents a novel approach of stability analysis as well as a new kind of pheromone deposition rule which outperforms the traditional approach of pheromone deposition used so far in all variants of ant system algorithms. Our future effort is focused in comparing the two kinds of deposition approach with other models of ant system like *M*ax-*M*in Ant System (***MMAS***) and Rank-Based Ant System and estimate the optimum parameter setting of proposed deposition approach for these models.